\documentclass[runningheads]{llncs}
\usepackage{amsmath}
\usepackage{amsfonts}

\usepackage[T1]{fontenc}
\usepackage{graphicx}
\usepackage{color}
\newcommand{\revision}[1]{{\color{black}#1}}
\newcommand{\revisioncr}[1]{{\color{black}#1}}

\newcommand{\ubold}[1]{\fontseries{b}\selectfont#1}

\usepackage[export]{adjustbox}
\usepackage{booktabs}

\newcommand{\interval}[1]{{\footnotesize\textpm#1}}

\begin{document}
\title{Using Foundation Models as Pseudo-Label Generators for Pre-Clinical 4D Cardiac CT Segmentation}
%
% If the paper title is too long for the running head, you can set
% an abbreviated paper title here
%
\author{Anne-Marie Rickmann\inst{1}\orcidID{0000-0002-7432-0782} \and
        Stephanie L. Thorn \inst{3} \and
        Shawn S. Ahn \inst{5} \and      
        Supum Lee \inst{3} \and % ???
        Selen Uman \inst{6} \and
        Taras Lysyy \inst{7} \and %????
        Rachel Burns \inst{3} \and
        Nicole Guerrera \inst{3} \and
        Francis G. Spinale \inst{8} \and
        Jason A. Burdick \inst{6,9} \and
        Albert J. Sinusas \inst{1,2,3} \and
        James S. Duncan \inst{1,2,4}
        }
\authorrunning{A. Rickmann et al.}
\titlerunning{Using Foundation Models as Pseudo-Label Generators}

% First names are abbreviated in the running head.
% If there are more than two authors, 'et al.' is used.
%
\institute{
Department of Radiology and Biomedical Imaging, Yale University \email{anne-marie.rickmann@yale.edu}\\
\email{james.duncan@yale.edu}\and 
Department of Biomedical Engineering, Yale University \and 
Section of Cardiovascular Medicine, Department of Internal Medicine, Yale University \and
Department of Electrical Engineering, Yale University \and 
Department of Surgery, University of Pennsylvania \and
Department of Bioengineering, University of Pennsylvania \and
Washington University School of Medicine in St Louis \and
Department of Cell Biology \& Anatomy, University of South Carolina School of Medicine \and
Department of Chemical and Biological Engineering and BioFrontiers Institute, University of Colorado Boulder
}
\maketitle              % typeset the header of the contribution
\begin{abstract}
Cardiac image segmentation is an important step in many cardiac image analysis and modeling tasks such as motion tracking or simulations of cardiac mechanics. While deep learning has greatly advanced segmentation in clinical settings, there is limited work on pre-clinical imaging, notably in porcine models, which are often used due to their anatomical and physiological similarity to humans. However, differences between species create a domain shift that complicates direct model transfer from human to pig data.

Recently, foundation models trained on large human datasets have shown promise for robust medical image segmentation; yet their applicability to porcine data remains largely unexplored. In this work, we investigate whether foundation models can generate sufficiently accurate pseudo-labels for pig cardiac CT and propose a simple self-training approach to iteratively refine these labels. Our method requires no manually annotated pig data, relying instead on iterative updates to improve segmentation quality. We demonstrate that this self-training process not only enhances segmentation accuracy but also smooths out temporal inconsistencies across consecutive frames. Although our results are encouraging, there remains room for improvement, for example by incorporating more sophisticated self-training strategies and by exploring additional foundation models and other cardiac imaging technologies.

\end{abstract}
\section{Introduction}

Cardiovascular diseases remain a leading cause of morbidity and mortality worldwide, driving the need for accurate diagnostic tools and effective treatment planning. Cardiac CT, particularly 4D or 3D+time acquisitions, plays an important role by providing detailed spatiotemporal information on cardiac structure and function. 
To fully leverage these rich imaging datasets, accurate segmentation of cardiac structures is essential. 
However, manual segmentation is time-consuming and prone to inter- and intra-observer variability. These limitations have led to a surge in deep learning research for automated cardiac segmentation.
In recent years, foundation models, large-scale neural networks typically trained on extensive, often publicly available datasets, have emerged as a promising avenue for fast, accurate medical image analysis. 
There exist many publicly available cardiac datasets, which can be used to pre-train models and then adapt them to smaller, clinical datasets via techniques such as unsupervised domain adaptation or self-training.
However, pre-clinical research with animal models, e.g. pigs, faces a significant challenge. Large publicly available imaging datasets are rarely available for these species, and there is a substantial domain gap between human and animal cardiac images. This mismatch in anatomy, physiology, and imaging characteristics can degrade the performance of foundation models that were trained on human data alone.

To overcome this challenge, our work investigates whether and how foundation models trained on human data can be leveraged to generate initial segmentations for porcine cardiac CT. These noisy predictions then serve as pseudo-labels for a subsequent self-training process, where a deep learning model refines its own predictions over multiple iterations. 
A key aspect of this work is the emphasis on temporal consistency. In 4D cardiac imaging, consecutive frames capture the beating heart, and it is crucial for the segmentation to evolve smoothly from one frame to the next. Yet frame-by-frame segmentation, whether done manually or by deep learning, often suffers from small errors that manifest as discontinuities over time. We hypothesize that self-training can not only denoise the predictions but also enhance temporal stability.

In this paper, we make three main contributions. First, we demonstrate how foundation models (trained on human data) can still be used to generate meaningful pseudo-labels for porcine cardiac CT despite significant domain shifts. To the best of our knowledge, this is the first study to apply such models to pig data. Second, we propose a self-training strategy that iteratively refines these labels, potentially reducing noise and leading to improved segmentation performance. Finally, we present an evaluation of temporal consistency, showing that self-training using a frame-by-frame segmentation model can smooth out temporal inconsistencies. 

% old
\subsection{Related Work}
\noindent
\textit{Cardiac Image Segmentation:}
Deep learning approaches, particularly convolutional neural networks (CNNs) inspired by the U-Net architecture~\cite{ronneberger2015u}, have dominated medical image segmentation. Tools such as nnU-Net~\cite{isensee2021nnu} automate significant parts of the pipeline (e.g., data pre-processing, augmentation), facilitating rapid deployment on new datasets. Comprehensive overviews of deep learning in cardiac segmentation~\cite{chen2020deep,el2023deep} highlight how most methods rely on CNN-based architectures and emphasize the importance of improving both spatial and temporal coherence. Another key challenge is the limited availabilty of data and domain shift~\cite{el2023deep}.
Temporal consistency is critical for downstream tasks such as myocardial motion analysis. Various strategies have been proposed, including multi-task learning that combines segmentation and registration~\cite{Ta-multitask,yan2019cine} and temporal consistency losses~\cite{li2017fully}.
An alternative approach involves incorporating an additional temporal dimension into convolutional networks, as seen in 3D networks for 2D + time segmentation in echocardiography \cite{wei2020temporal,ling2023extraction}, and 4D CNNs for 3D + time cardiac CT data \cite{myronenko20204d}. However, 4D convolutions are not widely supported in deep learning frameworks and can lead to overfitting.
Another approach \cite{painchaud2022echocardiography}, involves a post-processing step that identifies and corrects temporal inconsistencies in segmentations using an autoencoder. 

\noindent
\textit{Learning from Limited Labels:}
Label scarcity is a well-known challenge in medical image analysis~\cite{cheplygina2019not}, particularly for segmentation. Semi-supervised and transfer learning methods aim to leverage large unlabeled datasets alongside smaller labeled subsets. In \emph{self-training}~\cite{bai2017semi,zou2018unsupervised,zou2019,shin2020,nie2018asdnet,yu2019uncertainty,xia20203d,strudel}, an initial model generates pseudo-labels for unlabeled data, which are then used to iteratively fine-tune the model. For a broader overview of methods for limited annotations, see~\cite{tajbakhsh2020embracing}.
\revision{
Traditionally, in self-training using pseudo labels, an initial model is trained on a small labeled dataset and then applied on unlabeled data to obtain pseudo labels. \cite{xie2020self}. The model is then retrained using a mixed dataset comprising both the labeled data and a subset of pseudo-labels that meet specific selection criteria (e.g., based on uncertainty or model confidence). This approach is necessary because during the early training phase, the model exhibits low accuracy and high entropy. The selective inclusion of pseudo-labels serves as a form of entropy minimization \cite{xie2020self,lee2013pseudo}. Xie et al. \cite{xie2020self} propose to have a separate teacher model and iteratively update the teacher with a trained student model, similar to our approach.}

\noindent
\textit{Segmentation Foundation Models:}
Foundation models, typically trained on large-scale data to be robust across domains, have gained attention in medical imaging~\cite{zhang2024challenges}. While vision-based foundation models such as SAM~\cite{kirillov2023segment} and SAM2~\cite{ravi2024sam} exhibit strong generalization, direct application to medical images often performs poorly without further adaptation. Modality-specific models such as TotalSegmentator~\cite{wasserthal2023totalsegmentator} (CT and MRI versions exist) also qualify as ``foundation'' in the sense that they are robust to unseen data and require minimal fine-tuning. Though some work has explored pseudo-labeling with foundation models~\cite{li2023segment,zhang2024improving,benigmim2024collaborating}, their application to non-human domains (e.g., pig data) and integration with self-training remain underexplored. 
\revision{
For example, Benigmim et al. \cite{benigmim2024collaborating} explore the use of foundation models for domain generalized semantic segmentation. They propose a collaboration of different foundation models, including using a text to vision model for generating additional training samples as data augmentation. Pseudo labels, generated by a fine-tuned CLIP model are further improved by using SAM.}
Relatively few studies focus on deep learning for pre-clinical porcine cardiac data, often due to the lack of large public datasets. One work uses transfer learning from \revision{models trained on human data~\cite{chen2018transfer}}, while another trains a U-Net directly on porcine cardiac MRI scans~\cite{kollmann2024cardiac}.

\section{Methods}
\begin{figure}
    \centering
    \includegraphics[width=\linewidth]{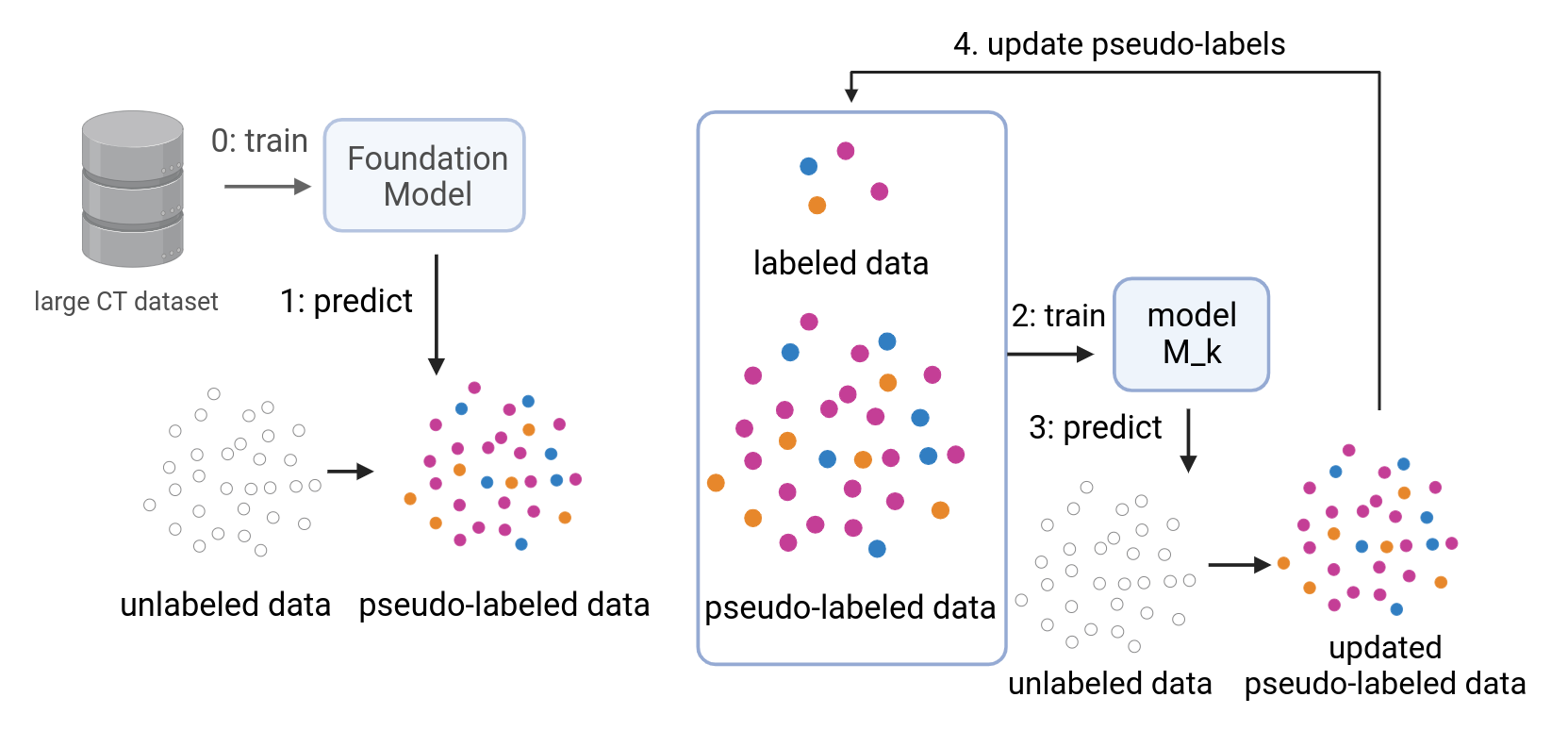}
    \caption{Self-training for 4D CT data, using a foundation model to initialize pseudo
labels, with an (optional) additional small labeled dataset. }
    \label{fig:self_training}
\end{figure}
\subsection{Methods}

To generate segmentation labels for porcine cine CT data, we employ an iterative self-training process initialized by a publicly available foundation model. Figure~\ref{fig:self_training} provides an overview of this iterative process.

Let $\mathcal{D} = \{ I_n \}_{n=1}^N$ denote a dataset of $N$ unlabeled 4D cine CT sequences. Each sequence $I_n$ is represented as
\[
    I_n: \Omega \times \{1,\ldots,T\} \to \mathbb{R},
\]
where $\Omega \subset \mathbb{R}^3$ is the 3D spatial domain, and $T$ is the number of time frames (e.g., $T=10$ for one cardiac cycle). Our goal is to produce a segmentation function
\[
    \hat{L}_n: \Omega \times \{1,\ldots,T\} \to \{0,1,\ldots,K\},
\]
where $K$ is the number of anatomical structures (plus background).

Let $M_F$ be a pre-trained 3D segmentation model, which is applied to each time frame independently. Splitting the 4D volume $I_n$ into $T$ frames $I_n^t$, we obtain initial pseudo-labels:
\[
    \hat{L}_n^t = M_F(I_n^t) \quad \text{for} \; t = 1,\ldots,T.
\]

Next, a new model $M_k$ is trained on these pseudo-labeled frames. We use a combination of Dice and Cross Entropy loss to compare $M_k(I_n^t)$ to $\hat{L}_n^t$. After training, $M_k$ generates updated pseudo-labels, which replace the old ones:
\[
    \hat{L}_n^t \leftarrow M_k(I_n^t),
\]
and the process is repeated for $K$ iterations.

In our experiments, TotalSegmentator is used to initialize pseudo-labels, and nnU-Net, specifically the Residual nnU-Net-L in full 3D resolution mode, serves as $M_k$. However, this workflow is generic and can accommodate other foundation models or segmentation architectures.

\section{Experiments \& Results}
\subsection{Data and Implementation Details}
\noindent
\textit{In-house porcine data}:
We created an in-house dataset from two porcine myocardial infarction (MI) studies.
In both studies reperfused MI was created by a 90 min balloon occlusion of the mid left anterior descending artery (LAD).  Imaging visits range from 1 to 6 visits over the span of up to 4 weeks. The data include images of healthy hearts and infarcts at various time points after reperfusion. Some pigs received intramyocardial injections of therapeutic hydrogels at 7 days post-MI using hyaluronic acid hydrogels.
All studies were approved by the Yale University School of Medicine Institutional Animal Care and Use Committee and according to the National Institute of Health Guidelines for Care and Use of Laboratory Animals.
This combined dataset contains 126 3D sequences from 28 pigs. Each sequence contains 8-10 frames, which leads to a total of 1249 frames. All scans were acquired during breath hold, so should not exhibit any motion artifacts due to breathing. Some scans have artifacts due to visible catheters in the LV or aorta.
We keep an additional smaller porcine dataset of 13 pigs with a total of 371 frames as a separate testing set. These scans include scans of pigs during thoracotomy, which was not seen in the model training.

\noindent
\textit{MMWHS}:
We use 20 3D CT scans from the training set of the MMWHS dataset~\cite{zhuang2016multi}, which consists of routine scans from healthy subjects.
This data is the only manually labeled dataset used in this work. We use it to validate our approach with ground truth labels. Further we use it to optionally mix in some manual labels into our training data.

\noindent
\textit{TotalSegmentator labels}:
For generating our pseudo labels, we use the TotalSegmentator~\cite{wasserthal2023totalsegmentator} heart chamber model, which segments the following labels: left ventricle (LV), left ventricle myocardium (LV myo), right ventricle (RV), left atrium (LA), right atrium (RA), aorta and pulmonary artery. 

\noindent
\textit{Implementation Details:}
We use the publicly available TotalSegmentator model for generating pseudo labels, and the publicly available nnU-Net code for training nnU-Net models. We follow the nnU-Net suggestions and use the Residual nnU-Net of size L, which requires a GPU with 24GB VRAM. We run all models on a cluster on A100 or A5000 GPUs.

\subsection{Results \& Discussion}
\noindent \paragraph{Preliminary Experiments with Human CT:} We first evaluated whether a self-training approach, initialized with pseudo labels generated by TotalSegmentator, can produce accurate segmentations on human CT data. Manually segmented scans were used as ground truth for validation, and the results are summarized in Table~\ref{tab:preliminary_mmwhs_transposed}.
We observe that TotalSegmentator predictions achieve high Dice scores (above 0.90) for most structures, with the exception of the aorta and pulmonary artery, both of which are difficult to segment. An nnU-Net model trained directly on manual labels achieves comparable performance and performs slightly better on these vessel structures. When trained exclusively on the pseudo labels, the model nearly matches TotalSegmentator’s performance, though it remains marginally lower for the aorta and pulmonary artery. Incorporating a small fraction of manual labels (5\% of the total) into the self-training process further boosts performance.
Since the dataset for this experiment was relatively small, we conducted a five-fold cross-validation for all models. Note that we only performed a single training iteration and did not update the pseudo labels.
\begin{table}[t]
\footnotesize
\center
\caption{Comparison of nnUNet models (ResEncM model) trained on the MMWHS dataset. All models were trained using 5 fold cross-validation. Manual: model was trained on the manual ground truth labels, TotalSegmentator: publicly available foundation model~\cite{wasserthal2023totalsegmentator}, Pseudo: the model was trained on pseudo labels obtained by applying TotalSegmentator to the training set, Mixed: model was trained on a mixed dataset of 95$\%$ pseudo labels and 5$\%$ manual labels. We provide mean and standard deviations of Dice scores, 95th percentile of the Hausdorff distance in mm (HD 95) and average symmetric surface distance in mm (ASSD).}
\begin{adjustbox}{width=\textwidth}
\begin{tabular}{l cc cc cc cc}
\toprule
\multicolumn{1}{c}{\ubold{Structure}} 
& \multicolumn{2}{c}{\ubold{Manual}} 
& \multicolumn{2}{c}{\ubold{TotalSegmentator}} 
& \multicolumn{2}{c}{\ubold{Pseudo}} 
& \multicolumn{2}{c}{\ubold{Mixed}} 

\\
\cmidrule(lr){2-3} \cmidrule(lr){4-5} \cmidrule(lr){6-7} \cmidrule(lr){8-9}
 & \ubold{Dice $\uparrow$} & \ubold{HD 95 (mm)$\downarrow$}
 & \ubold{Dice $\uparrow$} & \ubold{HD 95 (mm)$\downarrow$}
 & \ubold{Dice $\uparrow$} & \ubold{HD 95 (mm)$\downarrow$}
 & \ubold{Dice $\uparrow$} & \ubold{HD 95 (mm)$\downarrow$} 
\\
\midrule

\ubold{LV myo} 
& \ubold{0.922 \interval{0.021}} & \ubold{2.041 \interval{0.439}}
& 0.912 \interval{0.021} & 2.383 \interval{1.171}
& 0.914 \interval{0.017} & 2.312 \interval{0.886}
& 0.919 \interval{0.018} & 2.138 \interval{0.639}

\\

\ubold{LV} 
& \ubold{0.940 \interval{0.030}} & \ubold{2.258 \interval{0.830}}
& 0.933 \interval{0.050} & 2.595 \interval{1.669}
& 0.932 \interval{0.037} & 2.505 \interval{1.352}
& 0.938 \interval{0.032} & 2.300 \interval{1.131}

\\

\ubold{RV}
& 0.908 \interval{0.035} & \ubold{4.431 \interval{2.462}}
& \ubold{0.909 \interval{0.035}} & 6.151 \interval{5.684}
& 0.904 \interval{0.037} & 6.939 \interval{6.506}
& 0.908 \interval{0.037} & 5.938 \interval{5.691}

\\

\ubold{LA}
& \ubold{0.939 \interval{0.031}} & 4.171\interval{2.787}
& 0.934 \interval{0.031} & \ubold{3.627 \interval{1.908}}
& 0.936 \interval{0.031} & 3.705 \interval{2.049}
& 0.938 \interval{0.033} & 3.783 \interval{2.378}

\\

\ubold{RA}
& 0.912 \interval{0.046} & \ubold{6.417 \interval{5.404}}
& 0.911 \interval{0.037} & 4.772 \interval{2.587}
& \ubold{0.914 \interval{0.039}} & 4.678 \interval{2.300}
& 0.912 \interval{0.040} & 6.131 \interval{4.333}

\\

\ubold{aorta}
& \ubold{0.935 \interval{0.150}} & \ubold{3.534 \interval{9.638}}
& 0.652 \interval{0.060} & 71.42 \interval{9.571}
& 0.642 \interval{0.068} & 75.02 \interval{9.646}
& 0.692 \interval{0.112} & 68.27 \interval{20.59}

\\

\ubold{pulm. art.}
& 0.865 \interval{0.128} & 14.47 \interval{12.98}
& \ubold{0.882 \interval{0.067}} & \ubold{13.62 \interval{11.02}}
& 0.867 \interval{0.075} & 16.52 \interval{11.46}
& 0.871 \interval{0.092} & 14.09 \interval{12.28}

\\

\bottomrule
\end{tabular}
\end{adjustbox}
\label{tab:preliminary_mmwhs_transposed}
\end{table}

\paragraph{Porcine Data:} To evaluate our approach on porcine data, we first generate pseudo-labels for all 1249 frames using TotalSegmentator. We then train an nnU-Net model for 100 epochs, based on preliminary experiments indicating near-convergence at around 100 epochs (training Dice exceeding 0.9). After this training round, we replace the pseudo-labels with the model's predictions and train a new model for another 100 epochs. We repeat this process for a total of five sequential rounds. The final model obtained in this purely pseudo-labeled setup is termed \emph{pseudo only}.
We also explore a \emph{pseudo mixed} variant, which follows the same iterative procedure but incorporates 20 manually labeled 3D scans from the MMWHS dataset as additional training data in each round. Note that these ground-truth labels remain fixed and are not updated during self-training.

Since ground-truth porcine segmentations are unavailable, we cannot compute standard metrics such as Dice or Hausdorff distance. Instead, we assess segmentation plausibility by calculating the number of connected components, volume, and surface area for each predicted label. Segmentations are flagged if their volumes deviate from the mean by more than two standard deviations or if a single label contains more than one connected component. We make an exception for the aorta and pulmonary artery, where thin vessel structures easily lead to multiple components, and such instances could be easily corrected post-hoc.

\begin{figure}[h]
    \centering
    \includegraphics[width=\linewidth]{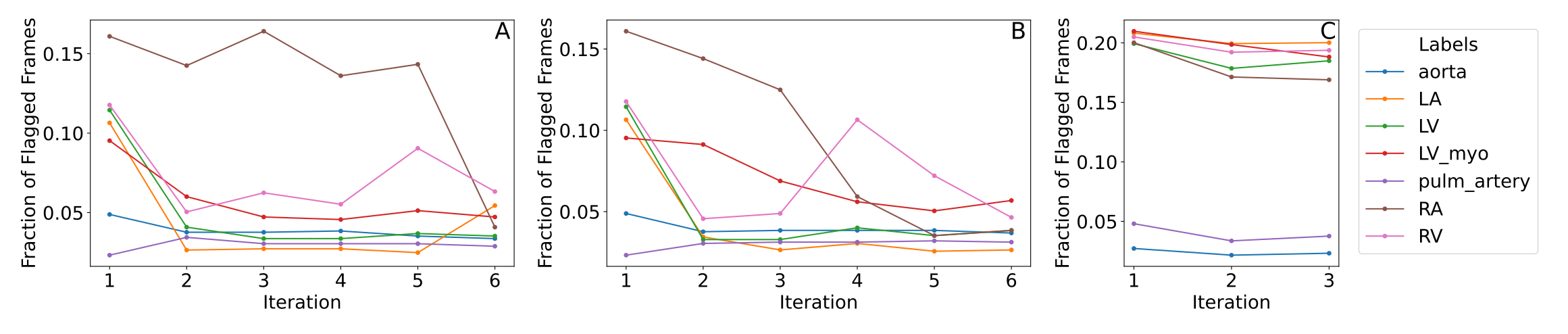}
    \caption{Fraction of flagged segmentations for each iteration, with iteration 1 = the initial foundation model predictions. A: The models were trained on TotalSegmentator pseudo labels only, B: The models were trained on a mix of TotalSegmentator pseudo labels and manually labeled human data. \revision{C: The models were trained in a standard self-training fashion where the model itself provides the initial pseudo labels.} }
    \label{fig:porcine_flag_plots}
\end{figure}

\begin{figure}[h!]
    \centering
    \includegraphics[width=0.95\linewidth]{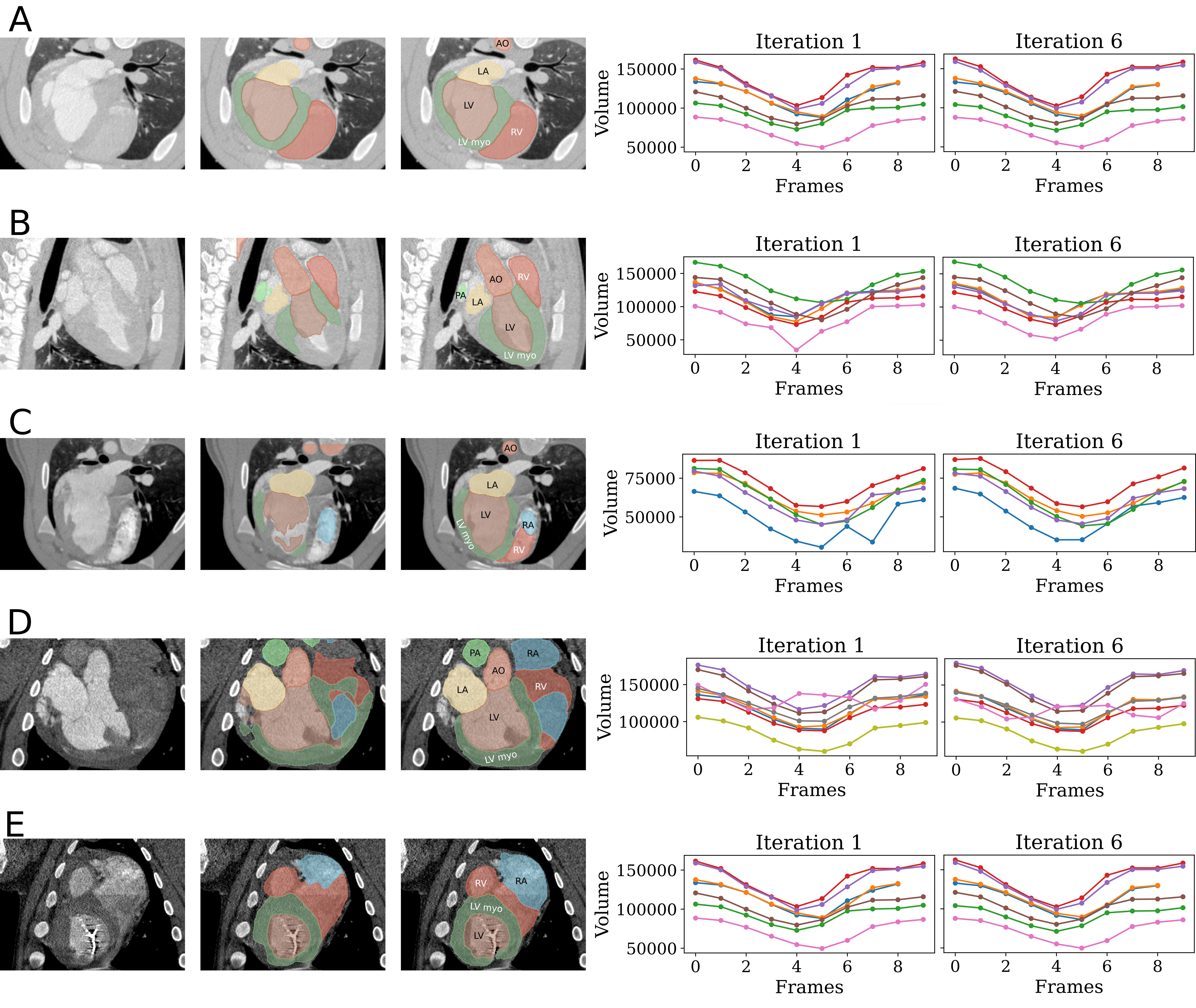}
    \caption{ Segmentation results \revision{using the "pseudo only" model} on porcine data, comparing the initial pseudo-labels (middle) and final pseudo-labels (right). \revision{Label abbreviations: RA: right atrium, LA: left atrium, RV: right ventricle, LV: left ventricle, LV myo: left ventricle myocardium, AO: aorta, PA: pulmonary artery}. The plots on the right show the temporal consistency of the left ventricle volume across frames, different colors represent different imaging visits of the same subject.
A: An example with an already accurate initial segmentation, resulting in minimal changes after self-training (frame 0 of the pink curve).
B: A case showing improvements through self-training (frame 4 of the pink curve).
C: Another case with large changes following self-training (frame 7 of the blue curve).
D: An example where slight improvements are achieved but some errors persist (frame 4 of the pink curve).
\revision{E: A case with an LV catheter that is causing artifacts. (frame 5 of the purple curve).}}
    \label{fig:segm_figure}
\end{figure}
To validate segmentation plausibility, we computed the percentage of frames flagged for each iteration (Figure~\ref{fig:porcine_flag_plots}). After 5 iterations around $3-6\%$ of frames remain flagged. Including manual human-labeled scans (\emph{pseudo mixed}) did not substantially improve performance overall. While right atrium segmentations improved slightly more quickly, the left ventricle myocardium deteriorated marginally. Notably, retaining only the largest connected component for each structure could further reduce these flagged instances. \revision{In panel C of Figure~\ref{fig:porcine_flag_plots}, we compare to a standard self-training approach, where a model was first trained on the MMWHS dataset (the same model as in Table ~\ref{tab:preliminary_mmwhs_transposed} Manual), and then used to initialize pseudo-labels. As expected, due to the domain shift between human and pig images, the pseudo label quality is worse than pseudo labels generated by the more robust foundation model TotalSegmentator. The simple self-training process does not improve the quality of the pseudo labels after 2 iterations, so we decided to stop the training. We believe additional selection criteria for updating the pseudo labels are needed in this case \cite{strudel,xie2020self,lee2013pseudo}.}

\revision{An additional contributing factor to segmentation errors in the porcine data may be the presence of cardiac catheters, an artifact potentially not seen in the model’s human training data, see last row of Figure~\ref{fig:segm_figure}}. 
Our results demonstrate that (i) a \revision{foundation model trained on human data} can reasonably handle porcine data, and (ii) self-training without any pig-specific ground truth can successfully refine the initial segmentations. Further improvements could stem from more sophisticated self-training strategies, such as incorporating uncertainty-based weighting or additional data augmentation, as well as from including a small set of manually labeled porcine scans and better aligning pig and human data, e.g., via image rotation.

Next, we investigated whether self-training smooths out temporal inconsistencies in the porcine segmentations. Specifically, we monitored the segmentation volume of each structure across consecutive frames and looked for abrupt “jumps” indicative of frame-to-frame errors. Figure~\ref{fig:segm_figure} provides example volume-time plots alongside corresponding segmentation visualizations (additional plots for all labels and subjects can be found in the supplementary material). Overall, we observe that self-training substantially reduces these abrupt jumps, yielding smoother volume trajectories and more consistent segmentations over time.
We believe that this reduction in temporal inconsistencies also arises from training the network on all image frames. Consecutive frames, which typically exhibit only small differences due to heart motion, effectively serve as a form of data augmentation within the same scan. 
\revisioncr{Further, the original publication of the foundation model \cite{wasserthal2023totalsegmentator} does not specify the number of CT scans the heart chamber model was trained on, nor whether it was trained on CT scans from different phases of the cardiac cycle.}
This could contribute to the initial temporal inconsistencies, which our self-training process is then able to mitigate.

\begin{figure}
    \centering
    \includegraphics[width=\linewidth]{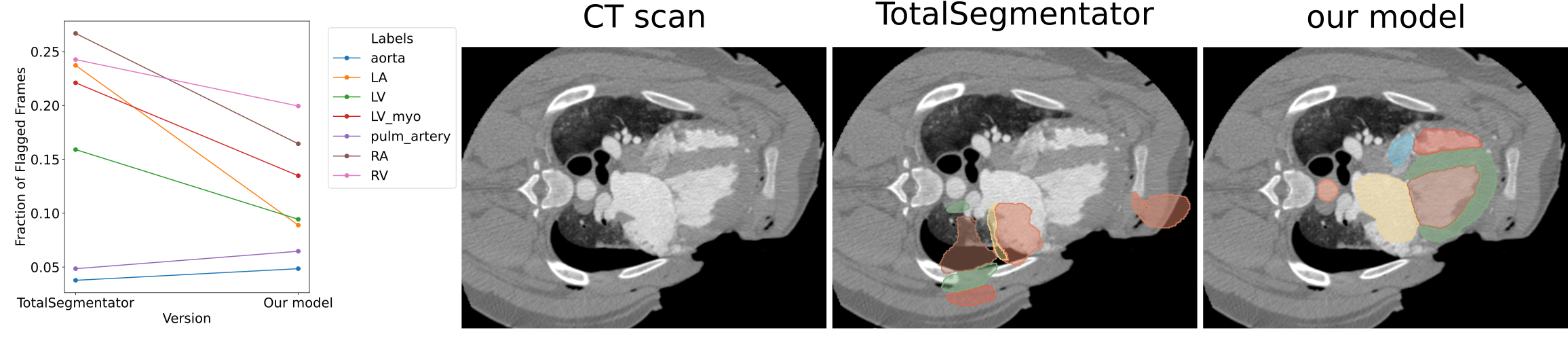}
    \caption{Fraction of flagged segmentations and example segmentation of TotalSegmentator and our \revision{"pseudo only"} model of unseen data.}
    \label{fig:unseen_data}
\end{figure}

Finally, we apply the trained model after 5 iterations to the unseen test set. This test set includes scans of pigs during thoracotomy, which was not seen in the training data. We show an example in Figure \ref{fig:unseen_data}. The fraction of flagged segmentations show that our model is more robust to unseen porcine data than TotalSegmentator.
\revisioncr{
To quantitatively assess temporal consistency beyond visual inspection, we computed two metrics. The standard deviation of Dice scores between consecutive cardiac frames and the average number of extreme points in a volume curve, similar to \cite{wei2023co}. We present those metrics for each anatomical structure in Table \ref{tab:temporal_consistency}. Lower values indicate more consistent segmentations across the cardiac cycle. Our iteratively refined approach achieved lower frame-to-frame Dice standard deviations and lower number of extreme points compared to the initial TotalSegmentator pseudo-labels, demonstrating improved temporal consistency.} The frame-to-frame Dice metric, while informative, does not account for differences due to natural cardiac motion, which could be addressed in future work through motion-compensated registration prior to evaluation.

\renewcommand\bfdefault{b}% rather than bx
\begin{table}[t]
\footnotesize
\setlength{\tabcolsep}{3pt} % Adjusted for better readability
\center
\caption{Comparison of temporal consistency between our refined segmentations and TotalSegmentator using two metrics: (1) the frame-to-frame Dice standard deviation and (2) the number of extreme points in volume curves. For each metric and anatomical structure, we report the mean and standard deviation across all subjects. Lower values for Dice standard deviation and extreme points indicate better temporal consistency.}
\begin{adjustbox}{width=\textwidth}
\begin{tabular}{l c c c c c c c}
\toprule
Method & \multicolumn{1}{c}{LV myo} & \multicolumn{1}{c}{LA} & \multicolumn{1}{c}{LV} & \multicolumn{1}{c}{RA} & \multicolumn{1}{c}{RV} & \multicolumn{1}{c}{aorta} & \multicolumn{1}{c}{pulm. art.}  \\
\midrule 
Ours (Dice) & 0.040 \interval{0.018} & 0.034 \interval{0.013} & 0.027 \interval{0.013} & 0.049 \interval{0.019} & 0.029 \interval{0.010} & 0.024 \interval{0.025} & 0.042 \interval{0.034}  \\
\addlinespace[0.3em]
TotalSeg (Dice) & 0.137 \interval{0.110} & 0.178 \interval{0.119} & 0.137 \interval{0.135} & 0.176 \interval{0.127} & 0.131 \interval{0.134} & 0.117 \interval{0.093} & 0.143 \interval{0.082}  \\
\midrule
Ours (Extremes) & 3.132 \interval{1.823} & 3.184 \interval{1.189} & 1.816 \interval{1.430} & 2.526 \interval{1.313} & 1.895 \interval{1.187} & 1.947 \interval{1.337} & 1.947 \interval{1.337}  \\
\addlinespace[0.3em]
TotalSeg (Extremes) & 4.158 \interval{1.405} & 3.526 \interval{1.272} & 2.526 \interval{1.666} & 3.421 \interval{1.330} & 3.500 \interval{1.446} & 3.605 \interval{1.288} & 3.184 \interval{1.393}  \\
\bottomrule
\end{tabular}
\end{adjustbox}
\label{tab:temporal_consistency}
\end{table}

\section{Conclusion} 
We investigated whether modality- and task-specific foundation models, can be leveraged to segment pre-clinical porcine images. Despite the notable domain shift between human and pig anatomy, our results show that the model’s initial predictions were reasonably accurate but required further refinement for practical use in pre-clinical research.
To address this gap, we explored a simple iterative self-training strategy in which the foundation model outputs serve as initial pseudo labels, and these labels are updated after every training cycle. Our findings indicate that this iterative process not only enhances overall segmentation quality but also mitigates frame-to-frame inconsistencies, \revision{likely due to training on multiple time-frames}.
Future work can build on this approach by incorporating more advanced self-training techniques and evaluating additional foundation models and imaging modalities. 
Such refinements may support more reliable and efficient cardiac imaging analyses in pre-clinical and translational research settings.

\section*{Acknowledgments}
This work was supported in part by the NIH grants R01HL121226,  R01HL175990, R01 HL170245, S10OD032277.

\section*{Disclosure of Interests}
The authors have no competing interests to declare that are relevant to the content of this article.

\bibliographystyle{splncs04}

\bibliography{references}

\end{document}